\documentclass[10pt,twocolumn,letterpaper]{article}

\usepackage{iccv}
\usepackage{times}
\usepackage{epsfig}
\usepackage{graphicx}
\usepackage{amsmath}
\usepackage{amssymb}

\usepackage{lipsum}
\usepackage{adjustbox}
\usepackage[accsupp]{axessibility}  %

\usepackage[pagebackref=true,breaklinks=true,letterpaper=true,colorlinks,bookmarks=false]{hyperref}

\iccvfinalcopy %

\def\iccvPaperID{8661} %

\ificcvfinal\fi

\begin{document}

\title{Boosting Monocular Depth Estimation with Lightweight 3D Point Fusion}

\author{Lam Huynh$^1$ \qquad
Phong Nguyen$^1$ \qquad
Jiri Matas$^2$ \qquad
Esa Rahtu$^3$ \qquad
Janne Heikkil\"a$^1$ \\
\small{$^1$University of Oulu \qquad $^2$Czech Technical University in Prague \qquad $^3$Tampere University} }

\maketitle
\ificcvfinal\thispagestyle{empty}\fi

\begin{abstract}
In this paper, we propose enhancing monocular depth estimation by adding 3D points as depth guidance. Unlike existing depth completion methods, our approach performs well on extremely sparse and unevenly distributed point clouds, which makes it agnostic to the source of the 3D points. We achieve this by introducing a novel multi-scale 3D point fusion network that is both lightweight and efficient. We demonstrate its versatility on two different depth estimation problems where the 3D points have been acquired with conventional structure-from-motion and LiDAR. In both cases, our network performs on par with state-of-the-art depth completion methods and achieves significantly higher accuracy when only a small number of points is used while being more compact in terms of the number of parameters. We show that our method outperforms some contemporary deep learning based multi-view stereo and structure-from-motion methods both in accuracy and in compactness.
\end{abstract}

\section{Introduction}

Depth estimation from 2D images is a classical computer vision problem that has been mostly tackled with methods from multiple view geometry~\cite{hartley2003multiple,szeliski2011structure}. Conventional stereo, structure-from-motion and SLAM approaches are already well-established and integrated to many practical applications. However, they rely on feature detection and matching that can be challenging especially when the scene lacks distinct details, and as a result the 3D reconstruction often becomes sparse and incomplete.

More recently, learning-based approaches have been introduced that enable dense depth estimation by exploiting priors learned from training images. In particular, monocular depth estimation that leverages only a single image together with learned priors has become a popular area of research, where deep neural networks are used to implement models that directly predict a depth map for given input image~\cite{ramamonjisoa2019sharpnet,chen2019structure,Hu2018RevisitingSI,liu2018planenet,Yin2019enforcing,huynh2020guiding}. While the basic idea is simple and attractive, the accuracy of the monocular depth estimation methods is limited by the lack of strong geometric constraints such as parallax. Thus, considerably more accurate depth maps can be achieved with deep learning based multi-view stereo methods~\cite{yao2018mvsnet,yao2019recurrent,Luo-VideoDepth-2020,bloesch2018codeslam}. However, the accuracy comes at the cost of increased computational complexity as multiple images need to be aggregated by the network to produce a single depth map. 

Another approach for dense depth estimation is to start from depth sensors like LiDARs, and use depth completion to interpolate the missing depth values based on RGB data. Despite of impressive results achieved by recent methods such as~\cite{zhang2018deep,park2020non,hu2020PENet} they are mainly suitable for cases with relatively high 3D point density, but do not perform well with sparse point clouds.

\begin{figure*}[!t]
\begin{center}
  \includegraphics[width=0.88\linewidth]{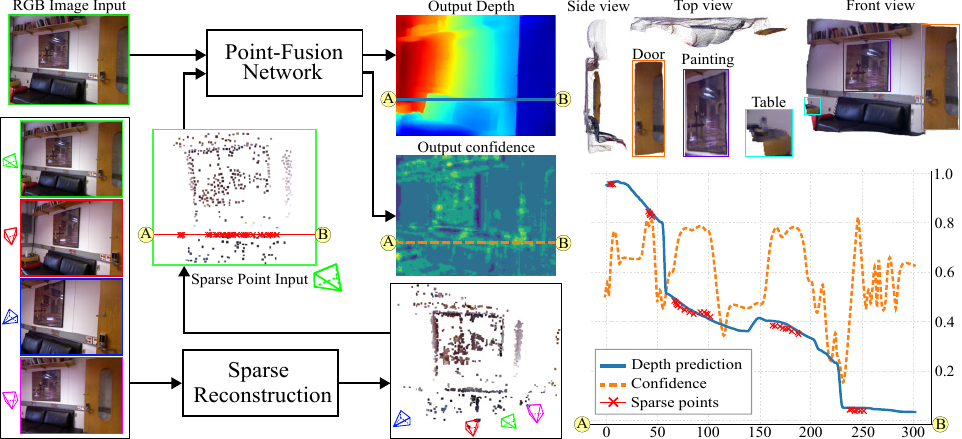}
\end{center}
  \caption{Dense depth prediction on the NYU-Depth-v2~\cite{silberman2012indoor} test set. A point cloud is produced by a conventional point-based sparse reconstruction method.
  The sparse 3D points and a single RGB image are 
  fed to the network  to estimate a high-quality depth map. 
  The dense reconstructed point cloud (top right) preserves the geometry well. The chart (bottom right) shows values
  along the A-B line of the estimated depth map, confidence map, and the sampling points. The estimated depth value tend towards sampling point values. Confidence value around sampling point areas are higher. (All values in the chart are normalized, and sparse point are enhanced for visualization)}
\label{fig:figure1_ver2}
\end{figure*}

In this paper, we start from monocular depth estimation and use a set of 3D points as constraints to obtain high-quality and dense depth maps as illustrated in Figure~\ref{fig:figure1_ver2}. The main difference of our approach to previous depth completion methods is that the point cloud can be extremely sparse and unevenly sampled, which enables using various methods for acquiring the 3D data, including conventional multi-view stereo, structure-from-motion and SLAM pipelines but also range sensors such as LiDARs. We argue that sparsity is important as it provides flexibility and cost savings to depth sensing. For example, in mobile imaging, existing AR frameworks, i.e. ARCore~\cite{google_arcore_2019}, ARKit~\cite{apple_arkit_2015}, and AREngine~\cite{huawei_arengine_2019}, provide sparse 3D point clouds, while in robotics and autonomous driving applications, low-resolution range sensors become sufficient. To this end, we propose a novel learning-based scheme for fusing RGB and 3D point data. More specifically, our contributions are the following:

\begin{itemize}
\item We introduce a novel multi-scale 3D point fusion neural network architecture, which is more lightweight than the existing state-of-the-art depth completion methods while being able to efficiently exploit the geometric constraints provided by a sparse set of 3D points.
\item We demonstrate state-of-the-art results on NYU-Depth-v2 and KITTI datasets with a network that uses only a fraction of the number of parameters compared to the other recent architectures.
\item We also show that our method combined with 3D point clouds obtained by using COLMAP~\cite{schoenberger2016mvs,schoenberger2016sfm} outperform recent deep learning based multi-view stereo and structure-from-motion methods both in accuracy and in compactness. 
\end{itemize}

\section{Related work}

\paragraph{\bf Single image depth estimation (SIDE):}
SIDE was first introduced by Saxena et al. \cite{saxena2006learning} and it gained momentum from the work by Eigen et al. \cite{eigen2014depth,eigen2015predicting}. Since then, the number of related studies has grown rapidly \cite{laina2016deeper,fu2018deep,qi2018geonet,ren2019deep,lee2019monocular,jiao2018look,Hu2018RevisitingSI,chen2019structure,facil2019cam,ramamonjisoa2019sharpnet,liu2018planenet,liu2019planercnn,lee2019big,huynh2020guiding}. At first, the proposed SIDE methods improved the accuracy by employing large architectures \cite{laina2016deeper,Hu2018RevisitingSI} and more complex encoding-decoding schemes \cite{chen2019structure}. Then, they started to diverge into using semantic labels \cite{jiao2018look}, exploiting the relationship between depth and surface normal \cite{qi2018geonet}, reformulating as a classification problem \cite{fu2018deep} or mixing both \cite{ren2019deep}. Other studies suggested to estimate relative depth \cite{lee2019monocular} or to learn calibration patterns to improve the generalization ability. Recent SIDE approaches exploit monocular priors such as occlusion \cite{ramamonjisoa2019sharpnet}, and planar structures either explicitly \cite{liu2018planenet,liu2019planercnn,Yin2019enforcing} or implicitly \cite{huynh2020guiding}. Despite these efforts, SIDE still generalizes quite poorly to unseen data. In this work, we leverage SIDE's ability to produce dense depth estimations and inject it with a small set of depth measurements to boost the accuracy while further shrinking the network size.

\paragraph{\bf Dense depth estimation from sparse depth:} Depth completion is a related problem where the aim is to densify or inpaint an incomplete depth map. Diebel and Thrun \cite{diebel2006application} is one of the first studies to tackle this problem using Markov random fields. Hawe et al. \cite{hawe2011dense} estimate disparity using wavelet analysis. The problem gained popularity as commodity depth sensors and laser scanners (or LiDARs) become more available. Uhrig et al. \cite{uhrig2017sparsity} proposed sparse convolution to train a sparse invariant network. Jaritz et al. \cite{jaritz2018sparse} leveraged semantics to train the network at varying sparsity levels. Ma et al. \cite{mal2018sparse} concatenated the sparse depth map to an RGB image, and used this RGBD volume for training. Xu et al. \cite{xu2019depth} filled in the missing depth values using the depth normal constraint. Imran et al. \cite{imran2019depth} addressed the depth completion problem using depth coefficients as a representation. Qiu et al. \cite{Qiu_2019_CVPR} suggested depth and normal fusion using learned attention maps. Methods based on a spatial propagation network (SPN) iterative optimize the dense depth map either in local \cite{cheng2018depth,cheng2019learning} or non-local \cite{park2020non} affinity. Chen et al. \cite{chen2019learning} suggested fusing features from an image and 3D points to produce the dense depth. However, these depth completion methods usually aim for outdoor environments and street views where the points come from a LiDAR.

\begin{figure*}[!t]
\begin{center}
  \includegraphics[width=0.9\linewidth]{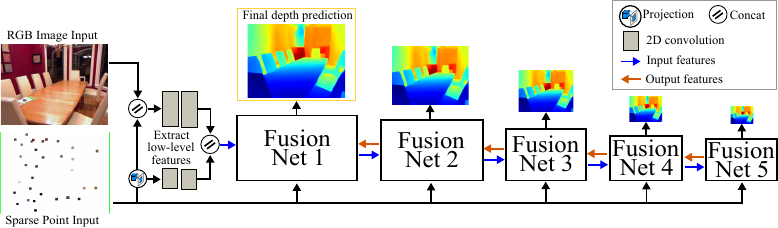}
\end{center}
  \vspace{-0.25cm}\caption{Overview architecture of the 3D point fusion network. Our model consists of five Fusion-Nets that iteratively extract and fuse 2D and 3D features at multiple-scale before predicting the final depth map at the highest spatial resolution.}
\label{fig:architecture_overview} \vspace{-0.15cm}
\end{figure*}

The difficulty of the depth completion problem much depends on the density of the 3D points used as an input to the algorithm. For example, LiDARs can produce relatively dense and regularly sampled point clouds without large holes, while passive image-based 3D reconstruction techniques, such as stereo or SLAM, result in substantially sparser set of points where the sampling is highly irregular and depends on the surface details. Thus, we argue that depth completion becomes a much harder problem when using a sparse point cloud from image-based reconstruction rather than from a LiDAR, and consequently, it also requires better regularization for the depth. To this end, we introduce a novel 3D fusion point network that efficiently learns to fuse image and geometric features to boost the performance of a monocular depth estimation network. It is a generic approach that can exploit RGB and 3D point data from various sources and environments. It can deal with indoor scenes that are often more diverse and challenging than outdoor environments, but it can be also used for depth estimation from street view scenes. Our work is inspired by \cite{chen2019learning}, but instead of sequentially fusing features at the same resolution, we build a deeper model to extract and fuse features at multiple-scales. This is crucial since \cite{chen2019learning} has been developed for depth completion of LiDAR data and as shown in our experiments it fails with a sparse set of points whereas thanks to the multi-scale approach our method can achieve reasonable accuracy from a few or even zero depth measurements.

\section{Method}
\begin{figure*}[!t]
\begin{center}
  \includegraphics[width=0.9\linewidth]{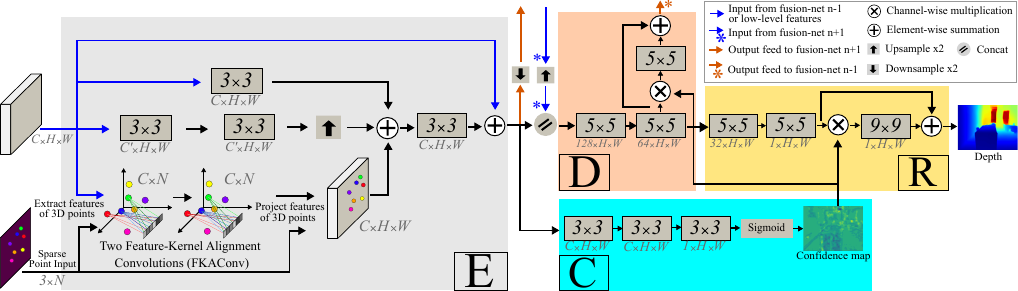}
\end{center}
  \caption{Details of  Fusion-Net $n \in[1, 5]$, where $n$ is the scale resolution. Main components include the feature fusion encoder (E), confidence predictor (C), decoder (D) and refinement (R) are color coded as gray, cyan, orange and yellow, respectively.}
\label{fig:architecture_components}
\end{figure*}

An overview of our 3D point fusion network is shown in Figure~\ref{fig:architecture_overview}. It is a fully convolutional framework that takes an RGB image and sparse 3D points as inputs to estimate a dense depth map. The 3D points serve as constraints to fix the overall geometry of the depth map produced by the network. To deal with the unstructured 3D point cloud, the points are first projected to the image plane and their $z$ coordinates are used to create a sparse depth map. Next, the RGB image is stacked with the sparse depth to form an RGBD image. We also apply two convolutional layers to the sparse depth and the RGBD image separately. The two outputs are concatenated to build the low-level input features that are fed to the first fusion-net module. The core network consists of five Fusion-Nets that operate at different feature resolutions. Each Fusion-Net contains a feature fusion encoder (E), a confidence predictor (C), a decoder (D), and a refinement (R) module as illustrates in Figure~\ref{fig:architecture_components}. We describe these modules in the following subsections and finish this section by giving details about our loss function.

\subsection{Feature Fusion Encoder}
Convolutional neural networks are good in processing regularly sampled data in a tensor form. Because our input point clouds are sparse and they represent geometric constraints unlike the image data, we cannot just rely on simple concatenation to fuse the information, but we need better representations. Inspired by a recent depth completion method~\cite{chen2019learning}, we design a feature fusion encoder to extract low-level features from RGB images and 3D points.

Our feature fusion encoder takes a 3D tensor ($C \times H \times W$) and a set of sparse points ($3 \times N$) as inputs, where $C$ is the number of feature channels, $H$ and $W$ are the height and width of the input tensor, and $N$ is the number of 3D points. The output is a 3D tensor with a similar shape to the input tensor. Details of the feature fusion encoder are shown in the gray box of Figure~\ref{fig:architecture_components}. It consists of two 2D branches, one 3D branch, and one convolutional layer for feature fusion. 

\noindent\textbf{The 2D convolutional branches:} The 2D branches are convolved at two different resolutions to learn multi-scale representations from the input 3D tensor. The first 2D branch has one convolutional layer with stride one to extract features at the same size as the input volume. The second 2D branch is a cascade of a stride two convolutional, a stride one convolutional, and an upsampling layer to obtain coarser features of the input tensor. The two outputs are summed to aggregate appearance features at different resolutions.

\noindent\textbf{The 3D point convolution branch:} The 3D branch aims to extract structural features from the sparse points. This is difficult for 2D convolutions that operate on local neighbors as 3D points are located on an irregular grid. Therefore, we utilize the feature-kernel alignment convolution (FKAConv)~\cite{boulch2020fka} that operates directly on 3D points to avoid this problem. The key idea of the FKAConv is to learn a linear transformation to align the neighboring points with the grid-like kernel. After that, it performs a weighted sum between this kernel and the features of the 3D points. One can see that 2D convolution is a special case where the learned linear transformation is always an identity matrix. 

As shown in Figure~\ref{fig:architecture_components}, our 3D branch consists of two FKAConv layers. We first extract the features of the 3D points from the input tensor using their projected 2D indices on the image plane. This volume has the size of $C \times N$. Next, we feed the point features and their 3D coordinates to the FKAConv layers. FKAConv selects a set of k-neighboring points for every input point and learns a transformation matrix to align the 3D points with its kernel. The point features are then convolved with the aligned 3D points to produce a 2D tensor of shape $C \times N$. The output features are projected back to an empty 3D tensor of size $C \times H \times W$ using the projected 2D indices. Features of other positions are set to zero.

\noindent\textbf{2D-3D Feature Fusion:} Output volumes from the 2D and 3D branches have the same shape as the input tensor ($C \times H \times W$). Therefore, to fuse these features, we sum them together before applying a 2D convolutional layer to output a 3D tensor of the size $C \times H \times W$. Finally, we add a residual connection to avoid vanishing gradient during training.

\subsection{Encoder, Decoder, and Confidence Predictor modules}
\noindent\textbf{Encoder and Decoder Module:} Designing efficient decoder and refinement modules is essential for the depth estimation problem~\cite{fang2020towards,wojna2019devil}. A common practice is to create large and complex decoders to produce accurate depth maps with sharp edges and fine details. However, we argue that by iteratively fusing relevant depth measurements from the 3D points with appearance features from image pixels, we can significantly reduce the size of our decoder and refinement designs. That is, our decoder and refinement modules have only two convolutional layers for each component. To simplify further, we use the same decoder and refinement designs for all Fusion-Nets. 

As shown in the orange box of Figure~\ref{fig:architecture_components}, the decoder transforms the fused features from the encoder before feeding them to the refinement module (the yellow box in Figure~\ref{fig:architecture_components}). We then initially obtain an output tensor of the decoder and a depth map. The estimated confidence map later modifies these two outputs.

\begin{figure}[!b]
\begin{center}
  \vspace{-0.25cm}\includegraphics[width=0.97\linewidth]{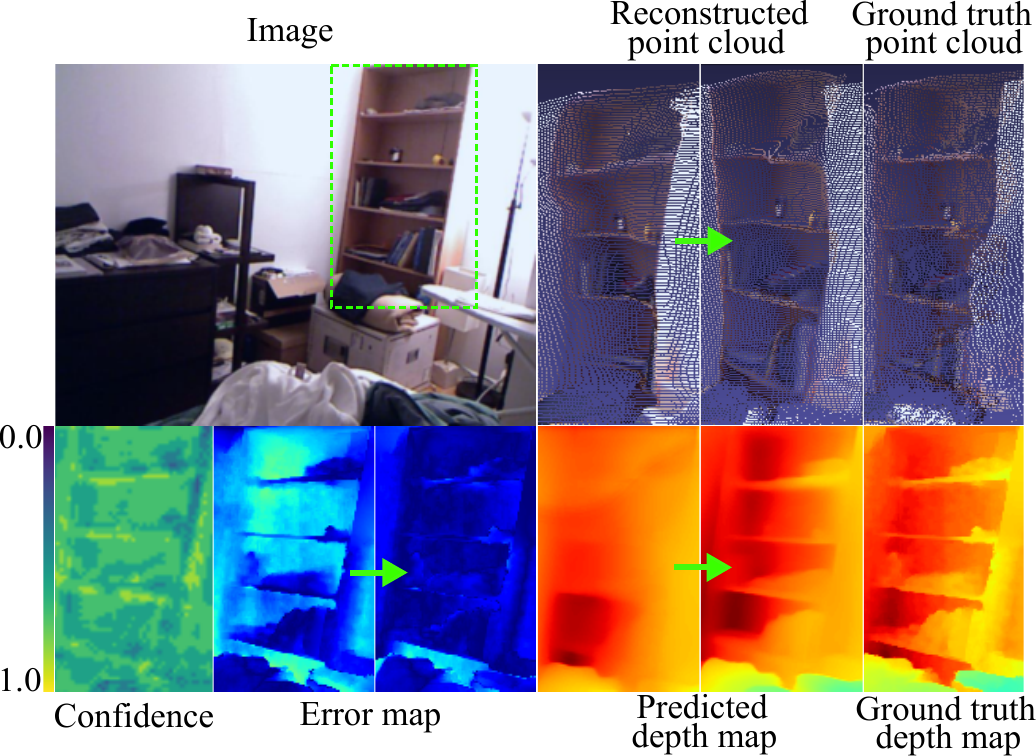}
\end{center}
  \vspace{-0.1cm}\caption{Depth map rectification by predicted confidence (green arrow). Confidence map values range from
  0.0 (low) to 1.0 (high). 
 }
\label{fig:confidence_map}
\end{figure}

\begin{table*}[h!]
\caption{\label{tab:eval_nyuv2}Evaluation on the NYU dataset. Metrics mark $\downarrow$ mean lower is better while $\uparrow$ is otherwise. Methods with $^{\ddagger}$ are trained using extra data. Figures \textit{with} $^{\star}$ indicates 3D COLMAP points while figures \textit{without} $^{\star}$
are obtained using randomly sampled points from GT depths.}
\centering
\small
\begin{tabular}{@{}llrrcccccc@{}}
\hline
\multicolumn{2}{c}{\textbf{Architecture}} &\hspace{-3ex}\textbf{\#3D pts}&\textbf{\#params} & \textbf{REL$\downarrow$} & \textbf{RMSE$\downarrow$} & \(\boldsymbol{\delta_{1}}\)$\uparrow$ & \(\boldsymbol{\delta_{2}}\)$\uparrow$ & \(\boldsymbol{\delta_{3}}\)$\uparrow$ \\ \hline
SharpNet & Ramam.'19$^{\ddagger}$~\cite{ramamonjisoa2019sharpnet} & 0 & 80.4M & 0.139 & 0.502 & 0.836 & 0.966 & 0.993 \\ \hline

Revisited mono-depth & Hu'19~\cite{Hu2018RevisitingSI} & 0 & 157.0M & 0.115 & 0.530 & 0.866 & 0.975 & 0.993 \\ \hline

SARPN & Chen'19~\cite{chen2019structure} & 0 & 210.3M & 0.111 & 0.514 & 0.878 & 0.977 & 0.994 \\ \hline

VNL & Yin'19~\cite{Yin2019enforcing} & 0 & 114.2M & \textbf{0.108} & 0.416 & 0.875 & 0.976 & 0.994 \\ \hline

DAV & Huynh'20~\cite{huynh2020guiding} & 0 & 25.1M & \textbf{0.108} & \textbf{0.412} & \textbf{0.882} & \textbf{0.980} & \textbf{0.996} \\ \hline 

\textbf{Point-Fusion} & \textbf{Ours} & 0 & \textbf{8.7M} & 0.128 & 0.505 & 0.847 & 0.971 & 0.994 \\ \hline \hline

NLSPN & Park'20~\cite{park2020non} & 2 & 25.8M & 0.300 & 1.152 & 0.393 & 0.697 & 0.879 \\ \hline
\textbf{Point-Fusion} & \textbf{Ours} & 2 & \textbf{8.7M} & \textbf{0.109} & \textbf{0.470} & \textbf{0.875} & \textbf{0.975} & \textbf{0.995} \\ \hline \hline

NLSPN & Park'20~\cite{park2020non} & 32 & 25.8M & 0.114 & 0.554 & 0.825 & 0.947 & 0.985 \\ \hline
\textbf{Point-Fusion} & \textbf{Ours} & 32 & \textbf{8.7M} & \textbf{0.057} & \textbf{0.319} & \textbf{0.963} & \textbf{0.992} & \textbf{0.998} \\ \hline \hline

Sparse \& Dense & Jaritz'18~\cite{jaritz2018sparse} & 200 & 58.3M & 0.050 & 0.194 & 0.930 & 0.960 & 0.991 \\ \hline

S2D & Ma'18~\cite{mal2018sparse} & 200 & 42.8M & 0.044 & 0.230 & 0.971 & 0.994 & 0.998 \\ \hline

GuideNet & Tang'20~\cite{tang2020learning} & 200 & 63.3M & 0.024 & 0.142 & 0.988 & 0.998 & \textbf{1.000} \\ \hline

NLSPN & Park'20~\cite{park2020non} & 200 & 25.8M & 0.019 & 0.136 & 0.989 & 0.998 & 0.999 \\ \hline

\textbf{Point-Fusion} & \textbf{Ours} & 200 & \textbf{8.7M} & \textbf{0.015} & \textbf{0.112} & \textbf{0.995} & \textbf{0.999} & \textbf{1.000} \\ \hline \hline

FuseNet & Chen'19~\cite{chen2019learning} & 500 & \textbf{1.9M} & 0.318 & 0.859 & 0.688 & 0.789 & 0.887 \\ \hline

CSPN & Cheng'18~\cite{cheng2018depth} & 500 & 18.5M & 0.016 & 0.117 & 0.992 & \textbf{0.999} & \textbf{1.000} \\ \hline

DeepLiDAR & Qiu'19~\cite{Qiu_2019_CVPR} & 500 &  53.4M & 0.022 & 0.115 & 0.993 & \textbf{0.999} & \textbf{1.000} \\ \hline

Depth Coefficients & Imran'19~\cite{imran2019depth} & 500 & 45.7M & 0.013 & 0.118 & 0.994 & \textbf{0.999} & - \\ \hline

DepthNormal & Xu'19~\cite{xu2019depth} & 500 &  29.1M & 0.018 & 0.112 & 0.995 & \textbf{0.999} & \textbf{1.000} \\ \hline

CSPN++ & Cheng'20~\cite{cheng2019learning} & 500 & 28.8M & - & 0.116 & - & - & - \\ \hline

GuideNet & Tang'20~\cite{tang2020learning} & 500 & 63.3M & 0.015 & 0.101 & 0.995 & \textbf{0.999} & \textbf{1.000} \\ \hline

NLSPN & Park'20~\cite{park2020non} & 500 & 25.8M & \textbf{0.012} & 0.092 & \textbf{0.996} & \textbf{0.999} & \textbf{1.000} \\ \hline

\textbf{Point-Fusion} & \textbf{Ours} & 500 & 8.7M & 0.014 & \textbf{0.090} & \textbf{0.996} & \textbf{0.999} & \textbf{1.000} \\ \hline \hline

MVSNet & Yao'18~\cite{yao2018mvsnet} & - & 124.5M & 0.043 & 0.162 & 0.940 & 0.972 & 0.996 \\ \hline

CodeSLAM & Bloesch'18~\cite{bloesch2018codeslam} & - & 66.3M & 0.096 & 0.251 & 0.910 & 0.962 & 0.989 \\ \hline

Consistent depth & Luo'20~\cite{Luo-VideoDepth-2020} & - & 178.2M & 0.086 & 0.345 & 0.916 & 0.959 & 0.984 \\ \hline

NLSPN & Park'20~\cite{park2020non} & 500$^{\star}$ & 25.8M & 0.042 & 0.144 & 0.949 & 0.981 & 0.999 \\ \hline

\textbf{Point-Fusion} & \textbf{Ours} & 500$^{\star}$ & \textbf{8.7M} & \textbf{0.022} & \textbf{0.126} & \textbf{0.994} & \textbf{0.999} & \textbf{1.000} \\ \hline \hline

\end{tabular} \vspace{-0.25cm}
\end{table*}

\noindent\textbf{Confidence Predictor:} Although the input 3D sparse points provide useful depth measurements, they can also contain noise. Hence, we proposed a simple yet efficient confidence predictor to attenuate the effect of noise. As illustrated in the cyan box of Figure~\ref{fig:architecture_components}, the output volumes from the feature fusion encoder are fed to three convolutional layers followed by a sigmoid to output the probability for every pixel. This information is then used to alter the initial depth map and the output features of the decoder. Moreover, we add residual connections at the end of the decoder and refinement blocks to prevent the vanishing gradient problem and regularize the confidence map's errors. The initial depth map is corrected based on the confidence map, as illustrated in Figure~\ref{fig:confidence_map}.

\subsection{Multi-scale Loss function}
We calculate the loss at multiple feature resolutions to train our network. The full loss is defined as:
\begin{equation} 
\label{eq:loss_total}
\mathcal{L} = \sum_{i=1}^{n=5} \gamma^{i} (\mathcal{L}^{i}_{log} + \mu \mathcal{L}^{i}_{grad} + \theta \mathcal{L}^{i}_{norm})
\end{equation}
where $n$ is the number of resolution scales and $\gamma^{i} \in \mathbb{R}^+$ is the loss weight at scale $i$, $\mathcal{L}_{log}$ is a variation of the $L_1$ norm that minimizes error on the sparse depth pixels, $\mathcal{L}_{grad}$ optimizes the error on edge structures, and $\mathcal{L}_{norm}$ penalizes angular error between the ground truth and predicted normal surfaces. These loss terms were introduced by Hu et al. ~\cite{Hu2018RevisitingSI} and widely adopted by state-of-art monocular depth estimation methods~\cite{chen2019structure,huynh2020guiding}. Subsection~\ref{implementation_detail} describes in detail how the network is trained using these loss functions.

\vspace{-0.15cm}
\section{Experiments}

In this section, we evaluate the performance of the proposed method and compare it with several baselines on the NYU-Depth-v2 and KITTI datasets. 

\subsection{Dataset and Evaluation metrics} 

\paragraph{\bf Datasets.} The NYU-Depth-v2 dataset contains approximately $120K$ RGB-D images recorded from 464 indoor scenes. We extract the raw RGB frames from the original videos and reconstruct sparse 3D point clouds using the COLMAP~\cite{schoenberger2016sfm,schoenberger2016mvs} structure-from-motion software. COLMAP is also used to extract the camera poses for multi-view stereo methods. The 3D points are back-projected to each input view to obtain a sparse set of depth values. We use 60K images for training and 654 images from the official test set for evaluating the methods. For KITTI, we utilize 85K images for training, 1000 images for validation and 1000 images for testing on the KITTI depth completion benchmark \cite{uhrig2017sparsity}.

\begin{figure}[!t]
\begin{center}
  \includegraphics[width=0.99\linewidth]{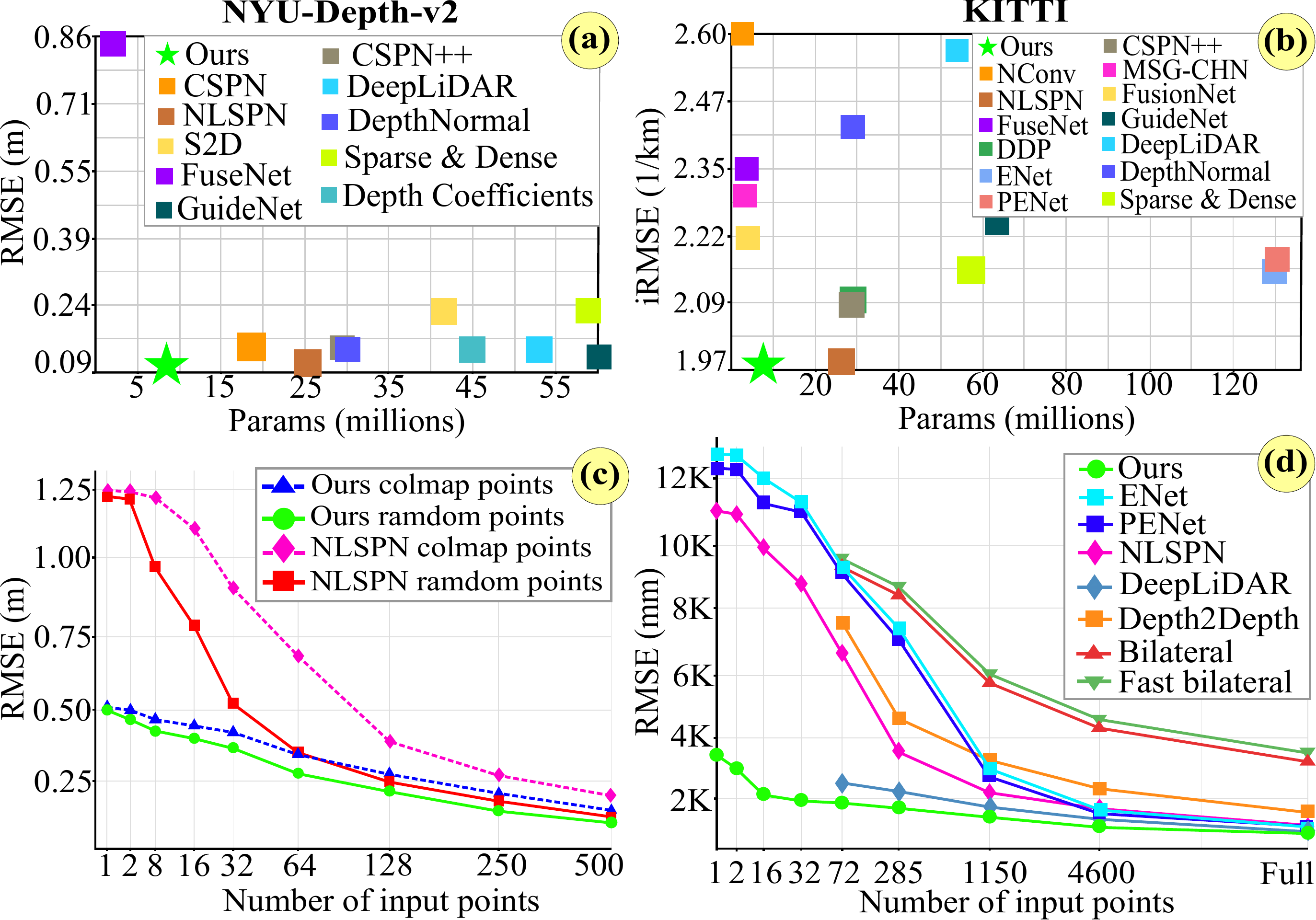}
\end{center}
  \caption{
  Top: RMSE and iRMSE metrics v. the number of parameters plots 
  for recent depth completion methods on NYU-Depth-v2 (a) and KITTI (b).
 Bottom: RMSE metrics for different sparsity and patterns
 for NYU-Depth-v2 (c) and KITTI (d).} \vspace{-0.25cm}
\label{fig:merged_chart}
\end{figure}

\vspace{-0.1cm}
\paragraph{\bf Evaluation metrics.} We report the results in terms of standard metrics for each dataset. For NYU-Depth-v2 we compute the mean absolute relative error (REL), root mean square error (RMSE), and thresholded accuracy ($\delta_i$). For KITTI, we also calculate RMSE plus mean absolute error (MAE), root mean square error (iRMSE) and mean absolute error (iMAE) of the inverse depth values. The detailed definitions of the measures are provided in the supplementary material. 

\begin{figure*}[!t]
\begin{center}
  \includegraphics[width=0.9\linewidth]{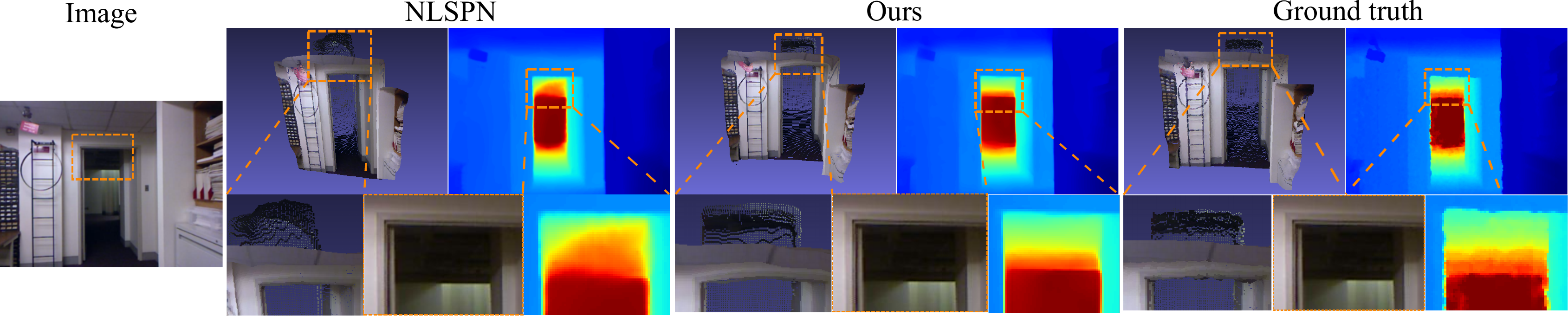}
\end{center}
  \caption{Qualitative results on NYU-v2 test set. Note that all methods use 200 randomly sampled 3D points as input.}
\label{fig:qualitative_nyu} \vspace{-0.5cm}
\end{figure*}

\subsection{Implementation details} \label{implementation_detail}

The proposed model is trained for 150 epochs on a single TITAN RTX using batch size of 32, the Adam optimizer \cite{kingma2014adam} with $(\beta_1, \beta_2, \epsilon) = (0.9, 0.999, 10^{-8})$, and the loss function presented in~(\ref{eq:loss_total}). The initial learning rate is $1.2*10^{-4}$, but from epoch 10 the learning is reduced by $6\%$ per $5$ epochs. We set the number of scales $n$ in~(\ref{eq:loss_total}) to 5, weight loss coefficients $\mu, \theta$ to $1.0$, and the scale weight losses $\gamma^1, \gamma^2, \gamma^3, \gamma^4, \gamma^5$ to $1.0, 0.75, 0.5, 0.25$ and $0.125$ respectively. To remove the effect of the arbitrary scale of the COLMAP points, we center and normalize the 3D inputs to a unit sphere before the training. During training, we augment the input RGB and ground truth depth images using random rotation ([-5.0, +5.0] degrees), horizontal flip, rectangular window droppings, and colorization (RGB only).

\subsection{Comparison with State-of-the-art}
The proposed method is related to multiple partially overlapping problems and, therefore, we compare it with several baseline methods in monocular depth estimation~\cite{ramamonjisoa2019sharpnet,Hu2018RevisitingSI,chen2019structure,Yin2019enforcing,huynh2020guiding}, depth completion~\cite{chen2019learning,cheng2018depth,cheng2019learning,eldesokey2019confidence,hu2020PENet,saif2021twise2,imran2019depth,jaritz2018sparse,li2020multi,mal2018sparse,park2020non,Qiu_2019_CVPR,tang2020learning,van2019sparse,xu2019depth,yang2019dense}, deep multi-view stereo~\cite{yao2018mvsnet}, deep structure-from-motion/SLAM~\cite{Luo-VideoDepth-2020,bloesch2018codeslam}. The baseline results are obtained using the pre-trained models \cite{chen2019structure,Hu2018RevisitingSI,ramamonjisoa2019sharpnet,Yin2019enforcing,Luo-VideoDepth-2020,hu2020PENet,eldesokey2019confidence}, re-training using the official NYU-v2~\cite{bloesch2018codeslam,mal2018sparse,park2020non,yao2018mvsnet} code, using our own re-implementations~\cite{huynh2020guiding,jaritz2018sparse}, and from the original papers \cite{cheng2019learning,Qiu_2019_CVPR,xu2019depth,imran2019depth,tang2020learning,yang2019dense}.

\paragraph{\bf NYU-Depth-v2.}

The performance metrics, computed between the estimated depth maps and the ground truth, are provided in Table \ref{tab:eval_nyuv2}. In addition, we report the number of method parameters, and the number of 3D points used in the estimation. Compared to monocular depth estimation studies, the proposed method provides a substantial improvement according to all metrics. For instance, REL, RMSE and thresholded accuracy ($\delta_i$) are improved by $47\%, 22.5\%$ and $10\%$, respectively, by using only $35 \%$ of the model parameters and 32 additional 3D points. Table~\ref{tab:eval_nyuv2} also shows that our method produces results close to state-of-the-art even without using any 3D inputs, while 2 points are already enough to be on par with the baseline approaches.

\begin{table}[t!]
\caption{\label{tab:eval_kitti}Evaluation results on the test set of the KITTI depth completion benchmark. Performance figures are color-coded as red, green, and blue, corresponding to first, second and third best results, respectively.}
\small
\adjustbox{max width=\columnwidth}{\begin{tabular}{l r @{\hspace{1.25\tabcolsep}} c @{\hspace{1.25\tabcolsep}} c @{\hspace{1.15\tabcolsep}} c @{\hspace{1.15\tabcolsep}} c}
\hline
\textbf{Architecture} & \textbf{\#param} & \textbf{RMSE} & \textbf{MAE} & \textbf{iRMSE} & \textbf{iMAE} \\ \hline

Sparse\&Dense~\cite{jaritz2018sparse} & 58.3M & 917.6 & 234.8 & 2.17 & 0.95 \\ \hline

NConv~\cite{eldesokey2019confidence} & \textcolor{red}{\textbf{0.36M}} & 829.9 & 233.2 & 2.60 & 1.03 \\ \hline

DepthNormal~\cite{xu2019depth} & 29.1M & 777.1 & 235.2 & 2.42 & 1.13 \\ \hline

FusionNet~\cite{van2019sparse} & 2.5M & 772.9 & 215.1 & 2.19 & 0.95 \\ \hline

FuseNet~\cite{chen2019learning} & \textcolor{blue}{\textbf{1.9M}} & 752.9 & 221.2 & 2.34 & 1.14 \\ \hline

DeepLiDAR~\cite{Qiu_2019_CVPR} & 53.4M & 748.4 & 226.5 & 2.56 & 1.15 \\ \hline

DDP~\cite{yang2019dense} & 29.1M & 832.9 & 203.9 & 2.10 & 0.85 \\ \hline

MSG-CHN~\cite{li2020multi} & \textcolor{green}{\textbf{1.25M}} & 762.2 & 220.4 & 2.30 & 0.98 \\ \hline

CSPN++~\cite{cheng2019learning} & 28.8M & 743.7 & 209.3 & \textcolor{blue}{\textbf{2.07}} & 0.90 \\ \hline

NLSPN~\cite{park2020non} & 25.8M & 741.7 & \textcolor{green}{\textbf{199.6}} & \textcolor{green}{\textbf{1.99}} & \textcolor{green}{\textbf{0.84}} \\ \hline

GuideNet~\cite{tang2020learning} & 63.3M & \textcolor{green}{\textbf{736.2}} & 218.8 & 2.25 & 0.99 \\ \hline

ENet~\cite{hu2020PENet} & 131.6M & \textcolor{blue}{\textbf{741.3}} & 216.3 & 2.14 & 0.95 \\ \hline

PENet~\cite{hu2020PENet} & 133.7M & \textcolor{red}{\textbf{730.1}} & 210.6 & 2.17 & 0.94 \\ \hline

TWISE\_2~\cite{saif2021twise2} & - & 840.2 & \textcolor{red}{\textbf{195.6}} & 2.08 & \textcolor{red}{\textbf{0.82}} \\ \hline

Point-Fusion(Ours) & 8.7M & 741.9 & \textcolor{blue}{\textbf{201.1}} & \textcolor{red}{\textbf{1.97}} & \textcolor{blue}{\textbf{0.85}} \\ \hline 
\end{tabular}}
\end{table}

\begin{figure}[!b]
\begin{center}
  \includegraphics[width=0.99\linewidth]{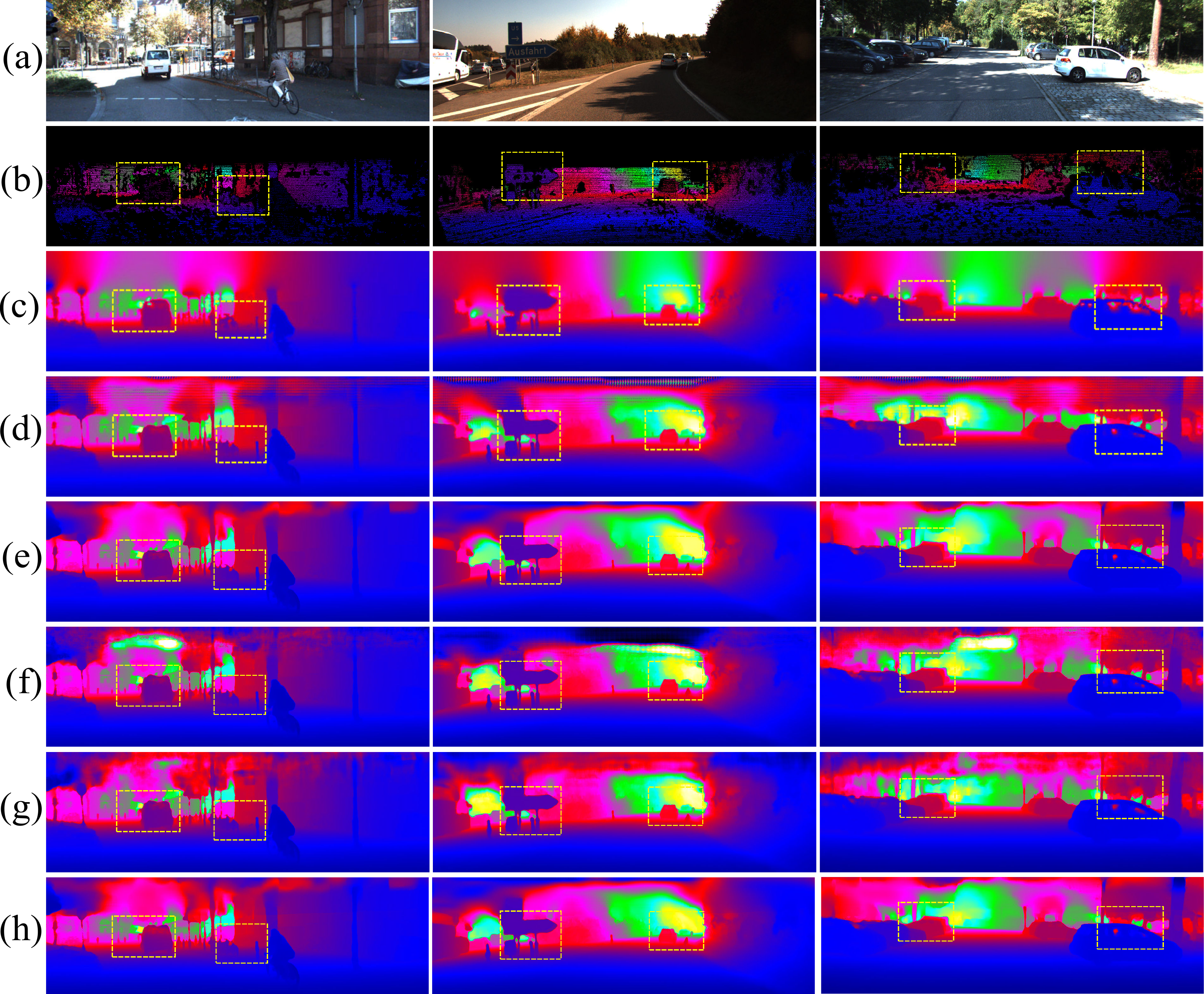}
\end{center}
  \caption{Examples from the KITTI validation set.
  Input images (a), ground truth LiDAR (b). Results
  from 
  (c) Depth2Depth~\cite{zhang2018deep},
  (d) DeepLiDAR~\cite{Qiu_2019_CVPR},
  (e) NLSPN~\cite{park2020non},
  (f) ENet~\cite{hu2020PENet},
  (g) PENet~\cite{hu2020PENet},
  and (h) the proposed method.
  }\vspace{-0.45cm}
\label{fig:qualitative_kitti} %
\end{figure}

Compared to the depth completion methods, we obtain state-of-the-art performance while using clearly less model parameters as shown in Figure~\ref{fig:merged_chart} a). 
The best performing baselines, NLSPN~\cite{park2020non}, DepthNormal~\cite{xu2019depth}, GuideNet~\cite{tang2020learning} use $2.9$, $3.4$, and $7.3$ times more parameters compared to our method, respectively. Instead of using the explicit 3D points, the multi-view stereo~\cite{yao2018mvsnet}, structure-from-motion~\cite{Luo-VideoDepth-2020}, and SLAM~\cite{bloesch2018codeslam} methods utilise multiple RGB images with camera poses. The results in Table \ref{tab:eval_nyuv2} indicate that the proposed model outperforms also these methods using only a fraction of the model parameters.

Figure~\ref{fig:qualitative_nyu} shows qualitative results of the predicted depth maps and reconstructed points cloud for our method and for~\cite{park2020non}. The baseline \cite{park2020non} results are obtained using the pre-trained model provided by the authors. Although both methods produce high quality depth maps, the proposed model is better in recovering fine details in challenging regions and introduces less distortions on flat surfaces. 

\begin{figure}[!b]
\begin{center}
  \includegraphics[width=0.81\linewidth]{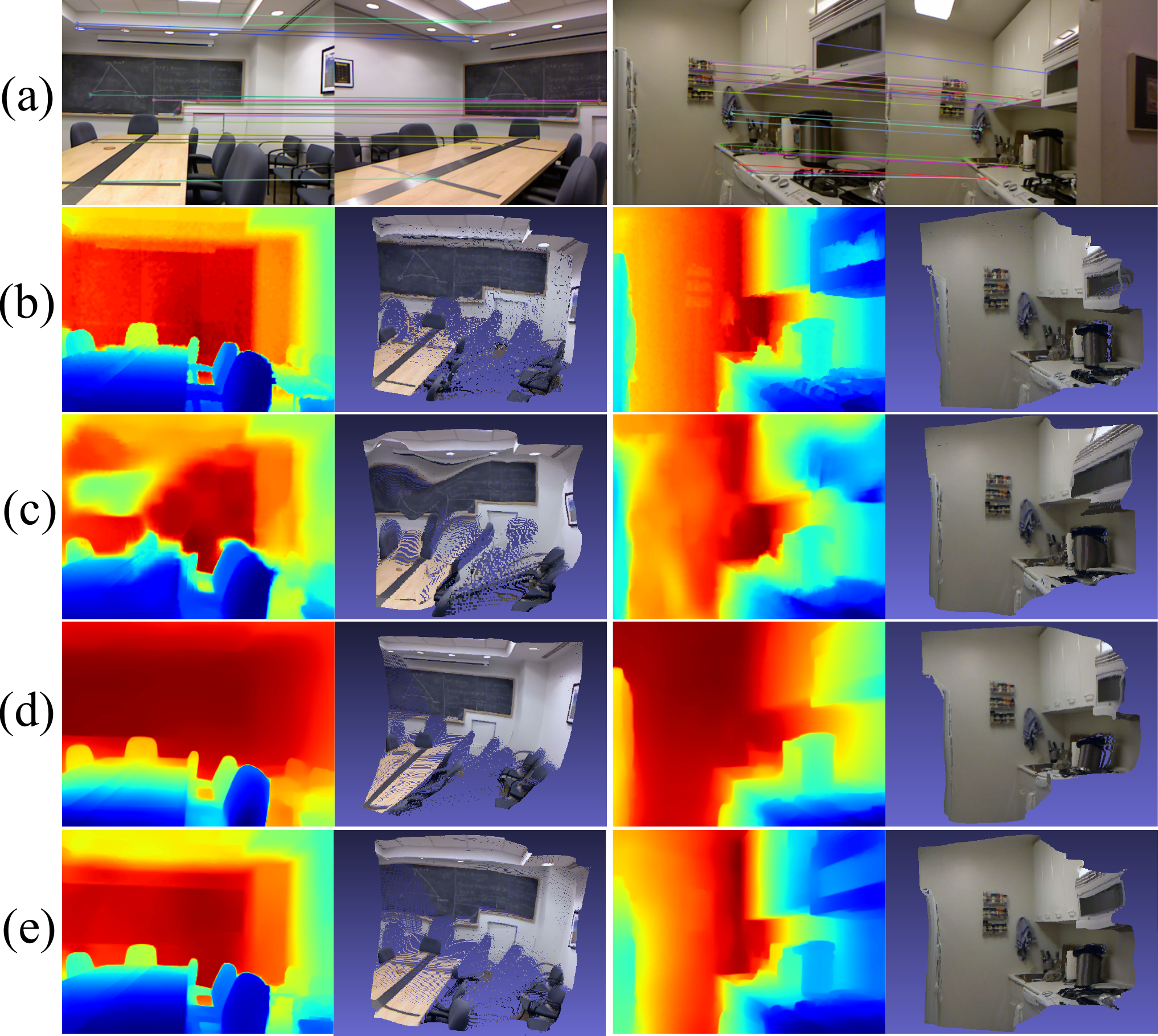}
\end{center}
\vspace{-0.1cm}
  \caption{
  NYU test set examples (a). Dense depth maps and reconstructed point cloud  from two images:   (b) ground truth, 
  (c) NLSPN~\cite{park2020non}, 
  (d) MVSNet~\cite{yao2018mvsnet}, 
  and (e) the proposed method.}
\label{fig:stereo_indoor}
\end{figure}

We also provide examples where we reconstructed a very sparse set of 3D points (32 points) from two images and utilized those as the 3D inputs. The dense depth maps obtained with this setting using our method, NLSPN~\cite{park2020non} and MVSNet~\cite{yao2018mvsnet} are illustrated in Figure~\ref{fig:stereo_indoor}.  We argue that state-of-the-art depth completion methods are usually vulnerable in high sparsity cases, while deep multi-view stereo performance degrades with less input views. One the other hand, our method produces high quality depth maps with significantly less distortions. Additional results are provided in the supplementary material.

\paragraph{\bf KITTI.}
To demonstrate the versatility of the proposed method, we also experiment with outdoor data.
For this purpose, we train and test our model with the KITTI depth completion dataset~\cite{uhrig2017sparsity}, in which we perform on par with state-of-the-art methods using a significantly smaller number of parameters as shown in Table~\ref{tab:eval_kitti}. We notice that there is a clear trade-off between model parameters and performance as shown in Figure~\ref{fig:merged_chart} b). Figure~\ref{fig:qualitative_kitti} presents qualitative comparison with baseline methods. The proposed method produces finer depth details as emphasized in the highlight areas. However, the difference is largest with a small number of input 3D points as depicted in Figure~\ref{fig:merged_chart} d). The results suggest that high-quality depth maps can be obtained by using only a few LiDAR points enabling more cost efficient solutions. Additional results are also added to the supplementary material.

\begin{figure*}[!t]
\begin{center}
  \includegraphics[width=0.93\linewidth]{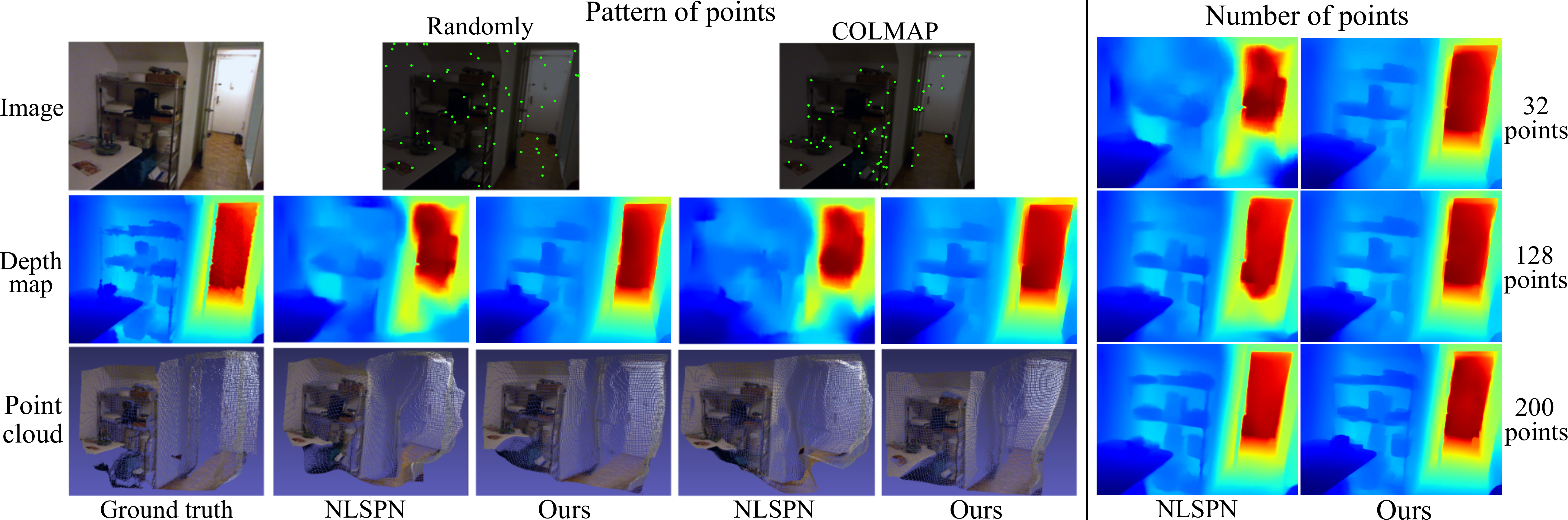}
\end{center}
  \vspace{-0.15cm}\caption{Qualitative comparison of the pattern (left) and quantity (right) of input points. Random 
 points are sampled from the dense ground truth depth map. For COLMAP points, we extract the image frames from raw NYU data and run COLMAP to obtain the points. The pattern and number of points are kept similar in all cases. Left: the number of points in use is 64. Results in the first row show the random input point have
 better spatial distribution than COLMAP points since they cover flat surfaces like walls, floors or doors. Right:  example results show the predicted depth maps using random sampled set of 32, 128 and 200 points respectively. Our method perform consistently better than NLSPN~\cite{park2020non} in all cases. (Points are enhanced for visualization)}
\label{fig:number_pattern_pts} \vspace{-0.45cm}
\end{figure*}

\subsection{Ablation studies} \label{ablation}

\paragraph{\bf Number and sampling of input 3D points.}
To analyze how the quantity and spatial distribution of the input 3D point affect the results, we performed experiments with varying 3D point patterns. For this purpose we generate sparse point sets by randomly sampling from the dense ground truth or from COLMAP output. We expect that by sampling from dense depth map provides better results compared to the COLMAP points. This is because, dense depth map covers also flat textureless surfaces such as walls, floor, and doors. However, such points might not be easy to obtain in practice, whereas COLMAP points represent location which are often reconstructed by SfM or SLAM methods. 

Figure~\ref{fig:merged_chart} c) presents the RMSE errors for different number of input points for both types. The results confirm the initial assumption that sampling from a dense depth map results in better performance. Moreover, we notice that the proposed method obtains higher accuracy compared to NLSPN~\cite{park2020non} with all point sets. In fact, we obtain similar performance using COLMAP points as NLSPN~\cite{park2020non} using points from the dense depth map. Figure~\ref{fig:number_pattern_pts} shows qualitative comparison with NLSPN~\cite{park2020non}.

\paragraph{\bf Confidence predictor.} We study the impact of the confidence predictor module by training our method with and without this component. We report the results in Table~\ref{tab:ablation_confidence_predictor}. When compared to a model without the confidence map, REL improves $\sim3.5\%$, and RMSE $\sim7.2\%$.

\begin{table}[b!]
\caption{\label{tab:ablation_confidence_predictor}Ablation studies of models without and with the confidence predictor (CP) on NYU-Depth-v2.}
\centering
\small
\begin{tabular}{lccccc}
\hline
\textbf{Training} & \textbf{REL$\downarrow$} & \textbf{RMSE$\downarrow$} & \(\boldsymbol{\delta_1}\)$\uparrow$ & \(\boldsymbol{\delta_2}\)$\uparrow$ & \(\boldsymbol{\delta_3}\)$\uparrow$ \\ \hline
w/o CP & 0.015 & 0.097 & 0.994 & 0.997 & 0.999 \\ \hline

\textbf{w/ CP} & \textbf{0.014} & \textbf{0.090} & \textbf{0.996} & \textbf{0.999} & \textbf{1.000} \\ \hline 
\end{tabular}
\end{table}

\paragraph{\bf Multi-scale Fusion-Net.} We assess how the number of Fusion-Nets affects the performance. For this purpose, we train our model using $2-6$ Fusion-Nets. The corresponding RMSE for the NYU-Depth-v2 test set are provided in Table~\ref{tab:ablation_fusionet}. The results improve by increasing the number of Fusion-Net to five and degrade after that. As each Fusion-Net perform at a different feature resolution, we argue that five is the optimal cascade size for the network to learn the geometric features from the 3D inputs.

\vspace{-0.5cm}
\begin{table}[b!]
\caption{\label{tab:ablation_fusionet}Performance of our model using different numbers of Fusion-Net on NYU-Depth-v2.}
\centering
\small
\begin{tabular}{l|c|c|c|c|c}
\hline
\# of FusionNet & 2 & 3 & 4 & \textbf{5} & 6 \\ \hline

RMSE & 0.105 & 0.097 & 0.094 & \textbf{0.090} & 0.093 \\ \hline
\end{tabular}
\end{table}

\paragraph{\bf 3D point convolutions.} We study the effect of different types of 3D point convolutions by training our model using the deep parametric continuous convolution (PCC)~\cite{wang2018deep} and the FKAConv~\cite{boulch2020fka}. The results are provided in Table~\ref{tab:ablation_point_conv}. The comparison with the PCC shows that the FKAConv module reduces the network size by $\sim5\%$ while slightly improves the performance by $\sim3\%$. We also trained our model without the 3D branch and the performance dropped considerably as shown in Table~\ref{tab:ablation_point_conv}.

\begin{table}[b!]
\caption{\label{tab:ablation_point_conv}Performance of our model applying different types of 3D point convolutions on NYU-Depth-v2.}
\centering
\small
\begin{tabular}{lrcccc}
\hline
\textbf{Training} & \hspace{-2ex}\textbf{\#params} & \textbf{REL$\downarrow$} & \textbf{RMSE$\downarrow$} & \(\boldsymbol{\delta_1}\)$\uparrow$ & \(\boldsymbol{\delta_2}\)$\uparrow$ \\ \hline
w/o 3D branch & \textbf{7.6M} & 0.044 & 0.196 & 0.980 & 0.993 \\ \hline

w/ PCC & 9.1M & 0.015 & 0.096 & 0.994 & 0.996 \\ \hline

\textbf{w/ FKAConv} & \textbf{8.7M} & \textbf{0.014} & \textbf{0.090} & \textbf{0.996} & \textbf{0.999} \\ \hline
\end{tabular}
\end{table}

\section{Conclusion}
We propose a novel and pragmatic approach that fuses RGB monocular depth estimation with information from a sparse set of 3D points for dense depth estimation. Experiments on common indoor and outdoor datasets show that we achieve state-of-the-art results while being compact in terms of the number of parameters. Moreover, our method can also produce, unlike the competitors, high-quality depth maps using an extremely sparse set of 3D points, which enables a cost-efficient solution for obtaining accurate depth maps for various applications where dense depth is needed.

\nocite{barron2016fast}

{\small
\bibliographystyle{ieee_fullname}
\bibliography{egbib}
}

\end{document}